# Estimating the time-lapse between medical insurance reimbursement with non-parametric regression models

Mary Akinyemi[1] and Chika Yinka-Banjo[1] and Ogban-Asuquo Ugot[1] and Akwarandu Nwachuku[1]

[1] University of Lagos, Akoka 100213 Lagos, Nigeria
cyinkabanjo@unilag.edu.ng

**Abstract.** Nonparametric supervised learning algorithms represent a succinct class of supervised learning algorithms where the learning parameters are highly flexible and whose values are directly dependent on the size of the training data. In this paper, we comparatively study the properties of four nonparametric algorithms, K-Nearest Neighbours (KNNs), Support Vector Machines (SVMs), Decision trees and Random forests. The supervised learning task is a regression estimate of the time lapse in medical insurance reimbursement. Our study is concerned precisely with how well each of the nonparametric regression models fits the training data. We quantify the goodness of fit using the R-squared metric. The results are presented with a focus on the effect of the size of the training data, the feature space dimension and hyperparameter optimization. The findings suggest k-NN's and SVM's algorithms as better models in predicting well-defined output labels (i.e, Time lapse in days). However, overall, the decision tree model performs better because it makes a better prediction on new data points than the ballpark estimates made from likelihood models- SVM's and k-NN's.

## 1 Introduction

Supervised learning in artificial intelligence is a relatively well-defined problem of estimating a target output label y, given an input vector x. Therefore, the goal of a supervised learning algorithm is to learn the objective function that maps a given input to an output, while minimizing a cost function. When the objective function to be learned by an algorithm has well defined fixed parameters such that the learning task is centered around estimating the values of these parameters, such an algorithm can be classified as a parametric algorithm, and thus will return a parametric model [12]. Parametric algorithms include; backpropagation, linear & logistic regression and naïve Bayes.

Nonparametric algorithms seek to best fit the training data through a “distribution” or (quasi) assumption-free model, whilst still maintaining some ability to generalize to unseen data. Nonparametric algorithms include; nearest neighbors (KNNs), decision trees and support vector machines (SVMs). We shall focus on the nonparametric algorithms and more specifically, nearest neighbors, support vector machines, decision trees and random forests. These algorithms have been extensively studied on diverse datasets [1], [2], [3]. We attempt to contribute to the comparative literature by studying the algorithms performance on a medical insurance dataset. The machine learning goal is centered around a regression model that estimates the time lapse between the date upon which medical treatment occurs and the date when an insurance company reimburses charges spent on the treatment.

The analysis focuses on the comparative effect of data preprocessing practices including, feature engineering and extraction, encoding and feature scaling, as well as other important aspects of machine learning like the hyper parameter tuning using the grid search algorithm. The $R^2$ score is chosen as the primary scoring technique simply because we wish to know how well each nonparametric regression model estimates the real data points, that is, the goodness of fit.

Section 2.0 of this paper presents a brief but precise literature review under the moniker related work, here we discuss some important comparative studies that make use of nonparametric algorithms in various applications. Section 3.0 presents our methodology for the comparisons, we also discuss the features of the dataset with important data analysis and visualization. Section 4.0 presents our results, section 5.0 discusses the results, and finally we conclude in section 6.0.

## 2 Related Work

The "no free lunch" theorem stated by David Wolpert [4], shows us that indeed, there is no one best supervised learning algorithm over another. What this means is that one can only say that an algorithm performs better in some task domains with a well-defined target label y and less so in other domains, this notion is key when carrying out comparative studies. The goal of a comparative analysis is not to state that one algorithm is better than the other but rather it is to reinforce certain expectations about the performance of an algorithm given a certain task. Most researchers are aware of the lack of a priori distinctions between supervised learning algorithms, and that the performance of an algorithm highly depends on the nature of the target label y. This is seen in the way the results from comparative studies are presented. In [5], the authors found that there is no universal best algorithm, however, using datasets from different substantive domains their results showed that boosted trees followed by SVMs had the best overall performance while logistic regression and decision trees performed the worst. Still based on the same prediction of recidivism, the results from [6], showed that when a subset of predictors was used, traditional techniques such as logistic regression performed better, while random forest performed the best on the whole set of predictors. More recent studies, have shown again, that the sample size is a key factor when comparing the performance of older algorithms such as logistic regression with newer algorithms like logitboost [7]. Research on breast cancer detection in [9] used five machine learning algorithms to separately classify a multidimensional image dataset and the results where compared. The five machine learning classifiers used on the Shearlton transformed images were, Support Vector Machines(SVM), Naive bayes, Multilayer perceptron, k-Nearest Neighbour, and Linear discriminant analysis classifier. The results conclude that SVM models were the best classifiers for breast cancer detection using images with a well-defined region of interest.

The study conducted in [8], compares Gradient Boosting Machine, Random Forests, Support Vector Machines, and Naive Bayes model. An ensemble of these models was fitted to create the ML model used for comparison with EUROSCORE II and logistic regression. The models were fitted twice due to the sensitivity to the input data, a chi-square fit was applied on the second fitting to the input features and only relevant features were used. The performance of each machine learning model was assessed with the area under the ROC curve, Random Forest produced the best result regardless if the data was filtered or not. Out of the four machine learning models, Naive Bayes produced the weakest accuracy without filtering, but proved to be better than SVMs with filtering; but in both cases logistic regression does better than both Naive Bayes and SVMs.
However brief, the key thing to take from the review is the various reasons why a particular learning algorithm performs better than another. While presenting our results, we shall discuss the performance of the algorithms with respect to certain aspects of the dataset as well as hyperparameter tuning. Although most of the papers reviewed where classification problems we are interested in seeing if similar effects of say the training set sample size as seen in [5] and [6] will affect the performance of our models.

## 3 Methodology

In this section, we present first the dataset used for the training of the models, then we present the various techniques used to train the models, test and score their performance.

### 3.1 The dataset

Choosing a health insurance plan is an important task that is often seldom paid as much attention as required. Studies have shown that most people rarely understand the concepts of cost-sharing, drug coverage and other benefits offered by insurance companies [10]. In Nigeria, Health Maintenance Organizations (HMOs) are often suggested to patients by hospitals, and these patients often lack detailed information as to the best HMOs to go with. With these challenges in mind, the dataset used in this paper was collected, with a view of uncovering, through exploratory analysis, possible information in health insurance records that could help with deciding the benefits of one HMO over another. The dataset is a health insurance record taken between the year 2015 and 2016. Collected, for a case study on Clearline HMO in Nigeria, the dataset contains 7 features, the treatment date, provider code (the hospital providing healthcare), diagnosis, drug prescription, charges spent (the cost of treatment), company code (the health insurance company) and the payment date (the date when the charges spent is reimbursed by the insurance company).

Our goal is to augment a statistical exploratory analysis of the data with supervised machine learning. The supervised learning task is to train a model to estimate the time lapse between a treatment date and payment date. To get the time lapse between reimbursement of the charges spent, we calculate the number of days between the treatment date and the payment date. This derived time-lapse is then used as the target variable during the training.

**Descriptive statistics.** In Table 1, we present a descriptive analysis of the health insurance dataset.

**Table 1.** Summary statistics for the numerical features of the dataset.

| S/N | Basic Statistics | Treatment date | Provider code | Charges sent | Company code | Payment date | Time lapse |
|---|---|---|---|---|---|---|---|
| 1 | Number of observations | 70888 | 70888 | 70888 | 70888 | 70888 | 70888 |
| 2 | Missing values | 0 | 0 | 0 | 0 | 0 | 0 |
| 3 | Mean | 2014.35 | 494.04 | 9980.69 | 196.27 | 2015.00 | 105.23 |
| 4 | Standard error of mean | 0.002 | 1.316 | 100.712 | 0.552 | 0.000 | 0.11 |
| 6. | Mode | 2015 | 580 | 5000 | 144 | 2015 | 106 |
| 7. | Standard deviation | 0.459 | 339.150 | 27419.072 | 148.730 | 0.000 | 0.013 |
| 8. | Variance | 0.210 | 111502.722 | 7.371E8 | 22120.487 | .000 | 0.00169 |
| 9. | Skewness | -1.415 | -0.083 | 13.344 | 0.441 | -0.023 | 15.33 |
| 10. | Kurtosis | 2.118 | -1.534 | 309.749 | -1350 | 1.33 | 405 |
| 11. | Standard error of kurtosis | 0.018 | 0.018 | 0.018 | 0.018 | 0.018 | 0.019 |
| 12. | Range | 5 | 1123 | 1346456 | 464 | 0 | 400 |
| 13. | Sum | 146408508 | 1294759 | 720396042 | 14249993 | 146428035 | 7469469 |
| 14. | Quantiles<br>25<br>50<br>75 | 2014.23<br>2014.74 | 106.15<br>580.06<br>782.86 | 2500.00<br>4330.00<br>8000.00 | 62.00<br>144.00<br>359.00 | 2015.58<br>2016.22 | 56.00<br>205.00<br>143.00 |

**Data Visualization.**

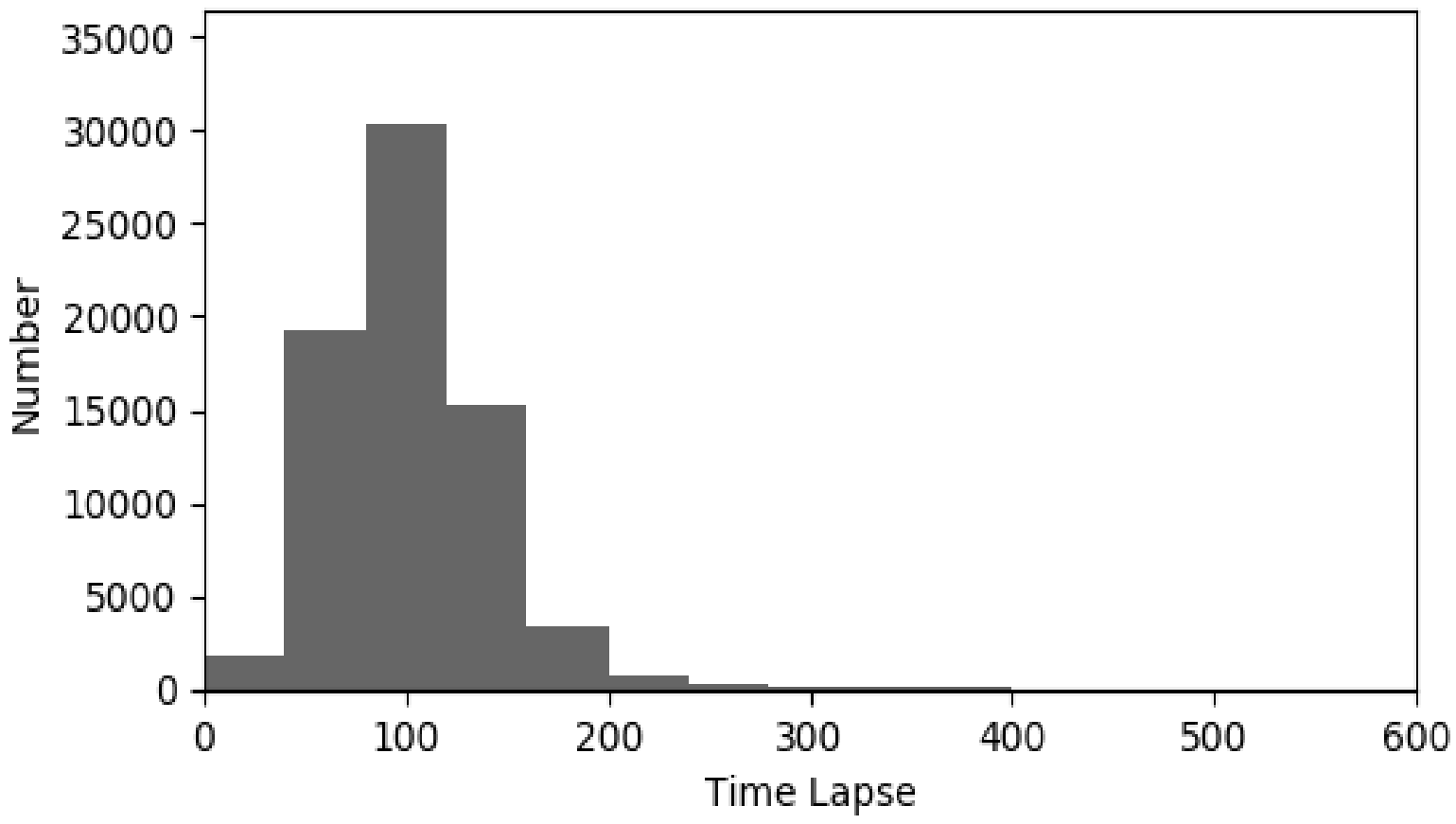

**Fig. 1.** Frequency distribution of the Time lapse

The frequency distribution in Fig. 1, shows that most insurance companies make reimbursements within 100 days.

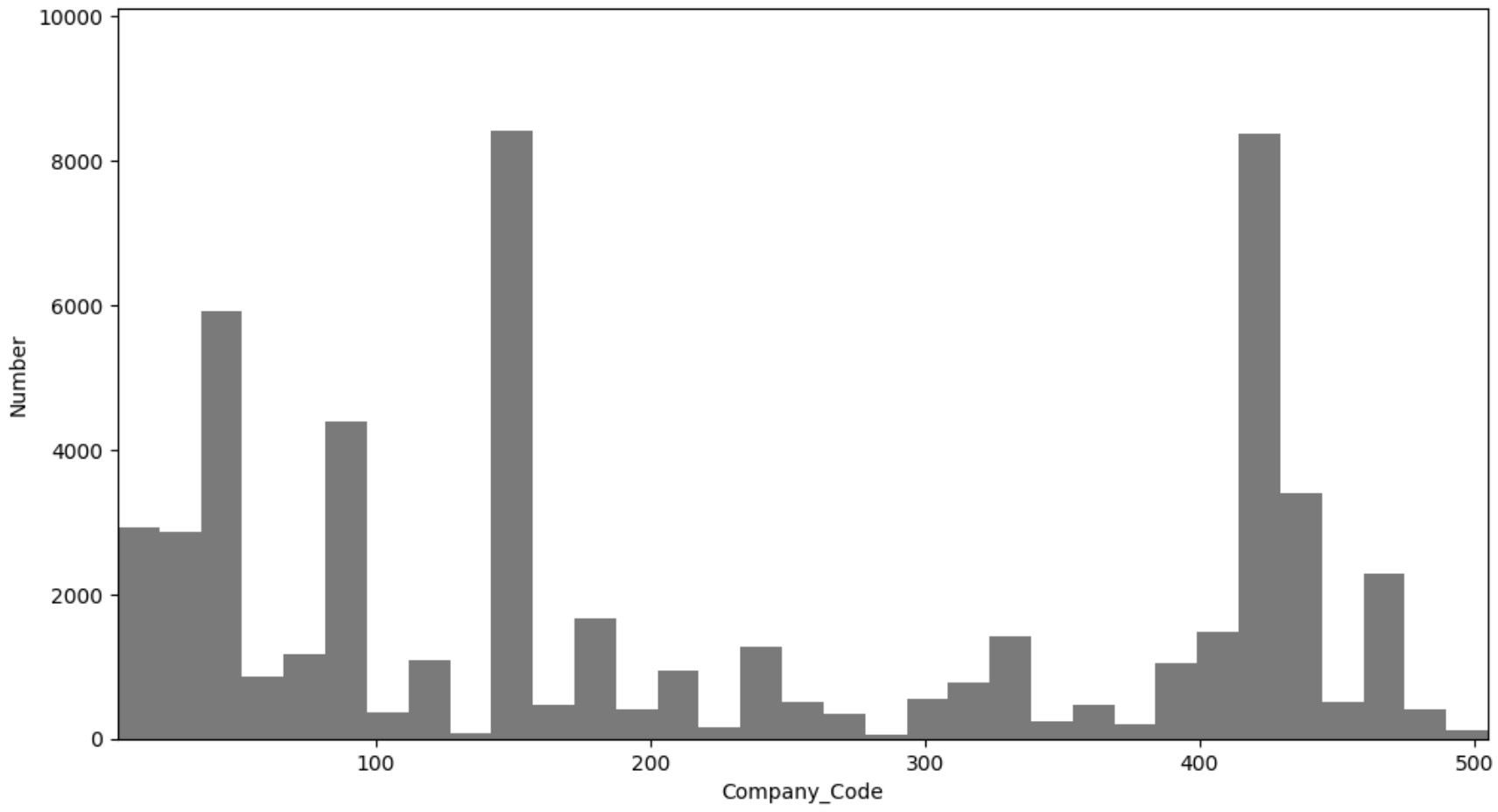

**Fig. 2.** Frequency distribution of the company code.

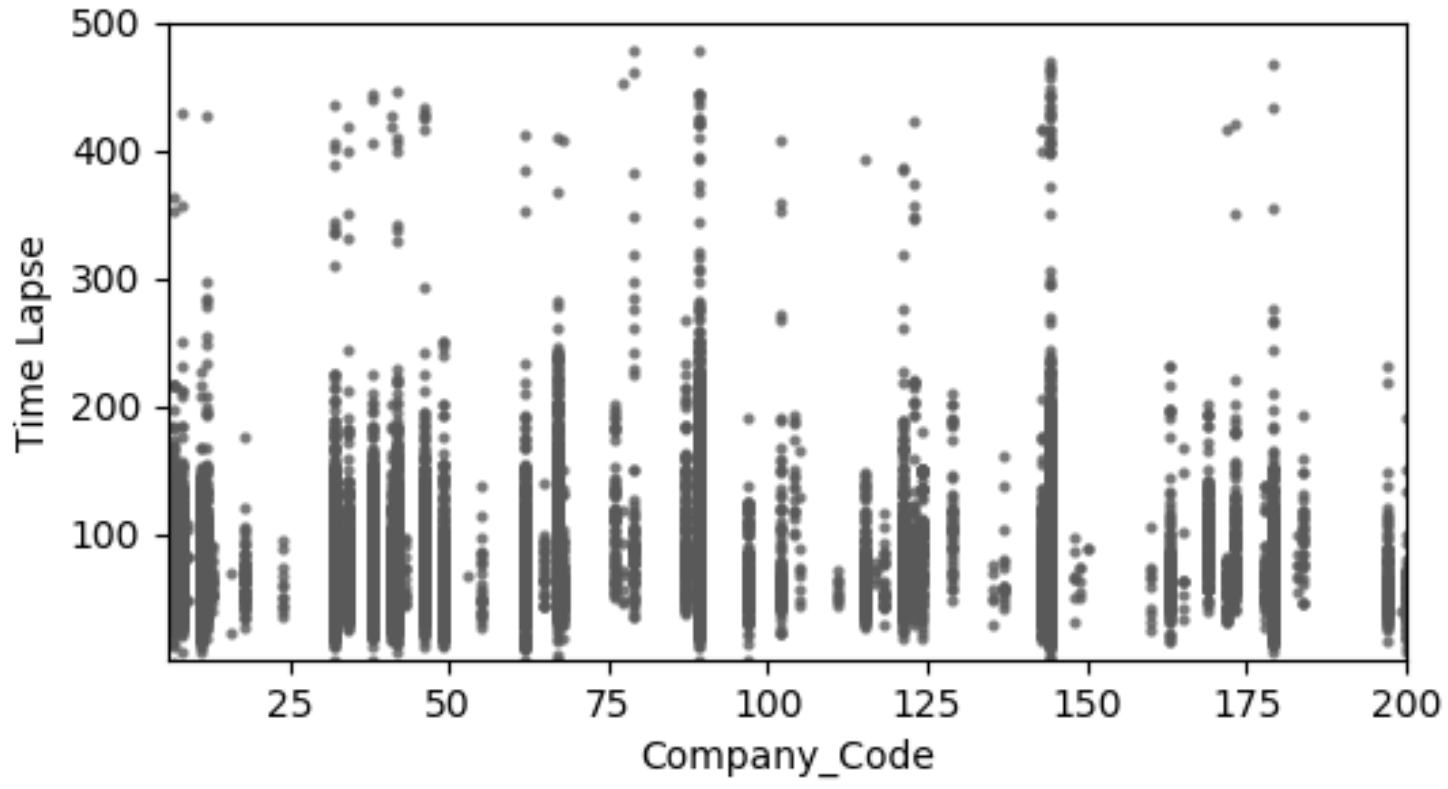

**Fig. 3.** Scatter plot showing the insurance companies (company_code) and the time lapse.

In Fig. 2 the frequency distribution of the company codes is presented, this indicates the number of records associated with each company. In Fig. 3, one can observe some insurance companies make reimbursements in less than 100 days while others may take as long as a year.

**Data preprocessing.** All the features except for the payment date are selected for our training and test set. The target label is of course the time lapse. We apply label encoding to the character features, the diagnosis and the prescription. Next, we apply one hot encoding to ensure that categorical features are not taken as ordinal values. This increased the number of features in the dataset to 533, we did not apply any dimensionality reduction techniques such as principal component analysis. We split the dataset into the training set and test set using the 80-20 (percent) ratio. This resulted in 56,709 instances for the training set and 14,178 instances for the test set
Finally, we normalize all the values in the training and test set to lie between 0 and 1 using the Minmax scaler function in sklearn. A mathematical description of the Minmax scaler is shown in equation 1, $x$ is an original value and $x'$ is the normalized value. Normalization ensures all values fall within a particular range and that an algorithm doesn't place a higher precedence on larger numerical values. The data preprocessing functions were also carried out using the sklearn library [11].

$$x' = \frac{x - \min(x)}{\max(x) - \min(x)} \quad (1)$$

### 3.2 Fitting the models

Every machine learning algorithm has a set of hyperparameters that determine to a very large extent how well the algorithm might perform, when fitting it to the data. Therefore, it was extremely important to choose the right hyperparameters for the models using an optimization technique. We rely on the Grid search algorithm to help us choose the best hyperparameters for each of the algorithms. Using grid search only solves half the problem actually, one still needs to know which hyperparameters to pass to the grid search algorithm for optimization. The grid search algorithm is fairly straight forward; given a set of hyperparameters α and β and a training set $S_{train}$, the goal of grid search is to find the optimal value w* such that:

$$w^*(\alpha, \beta) = argmin_w P(w, \alpha, \beta, S_{train}) \quad (2)$$

Where $P$ is the optimization problem of finding the optimal value of $w^*$. In equation 2, $w^*$is the model that minimizes the cost function and it is a function of the optimized parameters α and β. In Table 2 we present briefly the hyperparameters we chose for optimization, the range of optimization values and the effect of these choices.

**Table 2.** Hyperparameters chosen for optimization in the nonparametric algorithms.

| S/N | Hyperparameter | Algorithm | Range of values | Effect |
|---|---|---|---|---|
| 1. | K | KNN | 5, 10 | Lower k values may lead to high bias, higher k values may lead to high variance. |
| 2. | Tree search *algorithm* | KNN | k-D Tree, Ball Tree | Both are good for high dimensional datasets; ball tree has lower search time. |
| 3. | C | SVM | 0.1, 1 | It corresponds to regularize more the estimation. Lower values for noisy data. |
| 4.` | Kernel | SVM | Linear, RBF | Linear kernel works best for linearly separable data, RBF for non-linear data. |
| 5. | Max depth | Decision tree | 10, 20 | Shallow trees may lead to underfitting and deeper trees may lead to overfitting. |

### 3.3 Validation and scoring

To validate each model's performance on the training data, we employ the k-fold cross validation technique and set k to 10. The same k-fold cross validation is employed for the predictions on the test data as shown in Table 5.
To measure each model's goodness of fit, we use the $R^2$ metric. The $R^2$ metric is used simply because we only want to know precisely how well each of the models fit the data. The $R^2$ score ranges between 0 and 1, values closer to one are preferable and indicate a good fit. Equation 3 describes $R^2$:

$$R^2(y, y_{\text{actual}}) = \frac{\sum_{i=1}^{n-1}(y - \hat{y}) * (y - \hat{y})}{\sum_{i=1}^{n-1}(y - \underline{y}) * (y - \underline{y})} \quad (3)$$

Where $y_i$ is the estimated value of the $i$-th instance of $n$ samples, and $y_{\text{actual(i)}}$ is the actual $i$-th value of $n$ samples and $\underline{y} = \frac{1}{n}\sum_{i=0}^{n-1} y_i$.

## 4 Results

**Grid search results**

**Table 3.** Results from Grid search showing the optimized hyperparameters.

| S/N | Hyperparameters | Algorithm | Best value |
|---|---|---|---|
| 1. | K | KNN | 10 |
| 2. | Treesearch algorithm | KNN | Ball Tree |
| 3. | C | SVM | 1 |
| 4.` | Kernel | SVM | Linear |
| 5. | Max depth | Decision tree | 20 |

**Training results.**
The models are trained using the optimized parameters from table 3. Table 4 shows the training time for each model.

**Table 4.** Training time in seconds for each model.

| S/N | Algorithm | Training time (seconds) |
|---|---|---|
| 1. | KNN | 43 |
| 2. | SVM | 3367 |
| 3. | Decision tree | 548 |
| 4. | Random forest | 2395 |

**Validation results.**

**Table 5.** The mean of the 10-fold cross validation ($R^2$) score for each regression model.

| S/N | Regression model | $R^2$ |
|---|---|---|
| 1. | KNN | 0.4753 |
| 2. | SVM | 0.3631 |
| 3. | Decision tree | 0.6618 |
| 4.` | Random forest | 0.7189 |

**Test results.**

**Table 6.** $R^2$ score on the test data for each regression model.

| S/N | Regression model | $R^2$ |
|---|---|---|
| 1. | KNN | 0.4409 |
| 2. | SVM | 0.3403 |
| 3. | Decision tree | 0.6704 |
| 4.` | Random forest | 0.6813 |

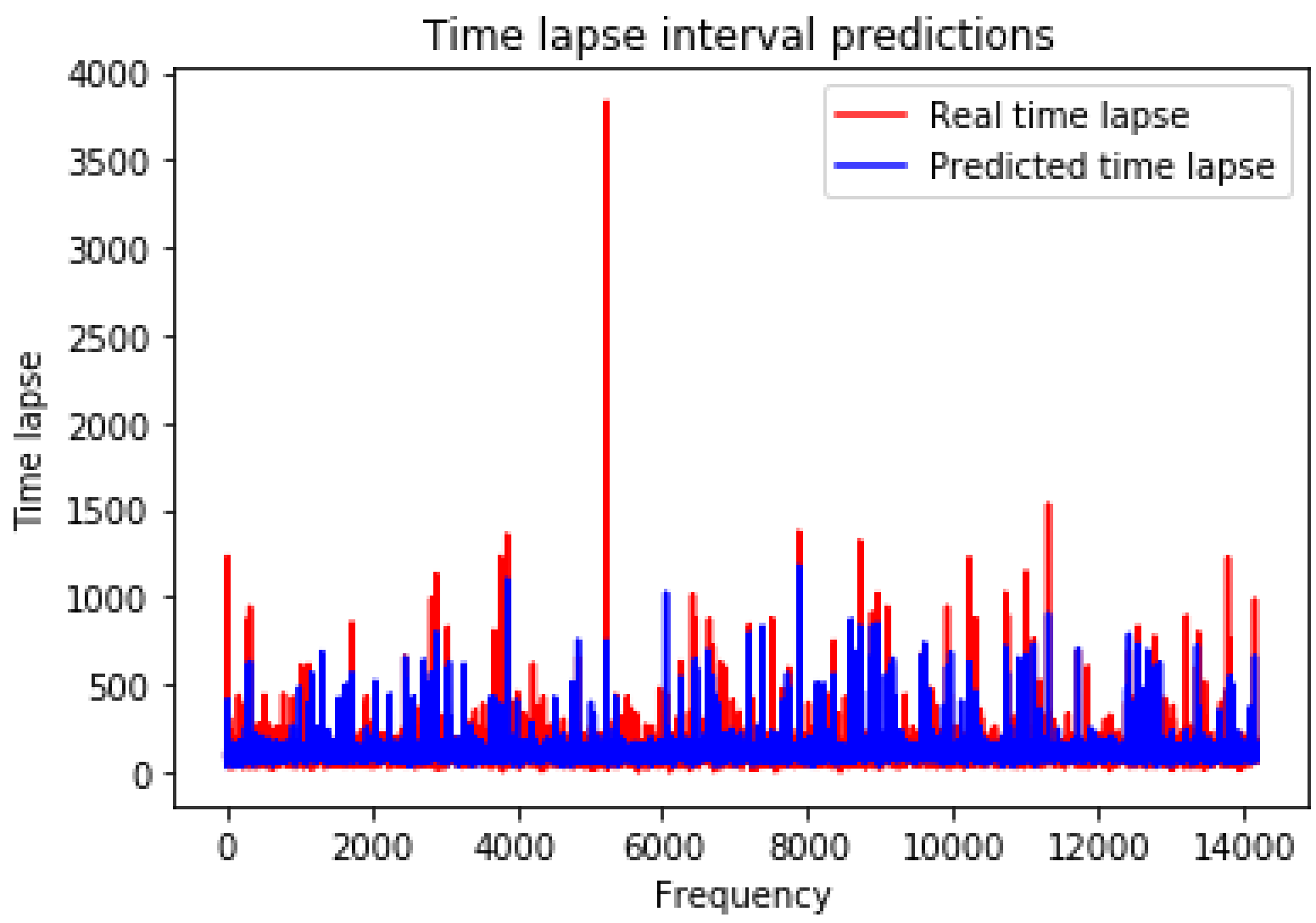

**Fig. 4.** Plot showing the predicted time lapse over the real time lapse for the KNN regression model.

The k-NN's visibly perform better than the SVM model. From Table 6, it is evident the K-NN model ascertains correct predictions on close to 50% of inputs, and finely predicts readings of a certain type of information in the data. The issue with this model is the rate of correct prediction on this type of information in the data is low. From the Fig. 4, in about 10 data points of input data, the model appears to correctly predict 3 of those data points correctly. This form of uncertainty brings the accuracy of the model down, although it performs better than SVM's it still does not top the work done by Decision trees or the Random forest model.

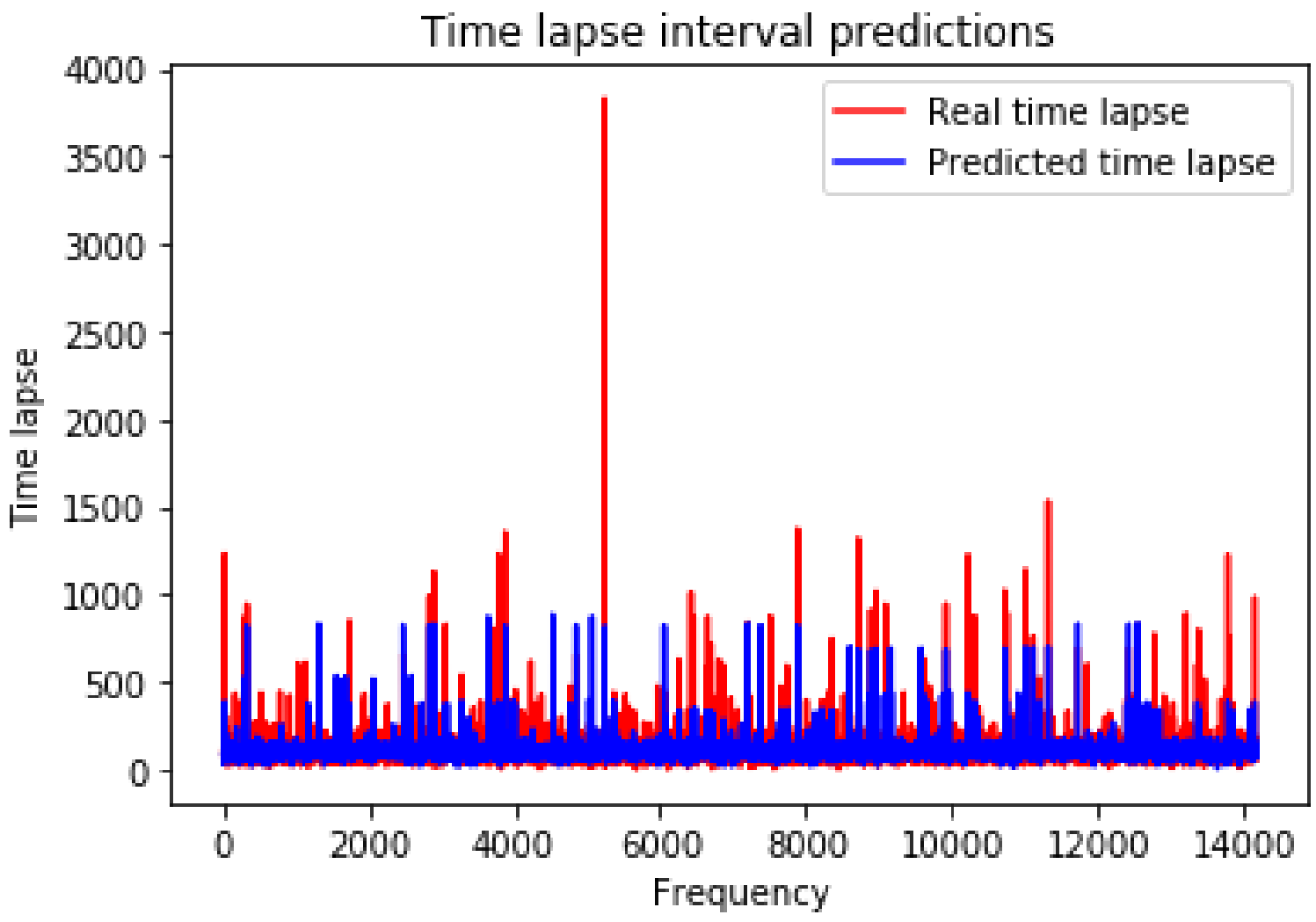

**Fig. 5.** Plot showing the predicted time lapse over the real time lapse for the SVM regression model.

The SVM's were not able to perform at a level of over 50% accuracy. This poor performance is observed in Fig. 5 and its $R^2$ shown in Table 6. The most common input readings (readings on the lower half of the plot), are not even correctly predicted by the model. On closer inspection, the SVM model could been deemed to have poorly underfit the data, as most of the new input data-points all fall into the same bracket or range of prediction. The nature of the data and the SVM model are not a good match at all.

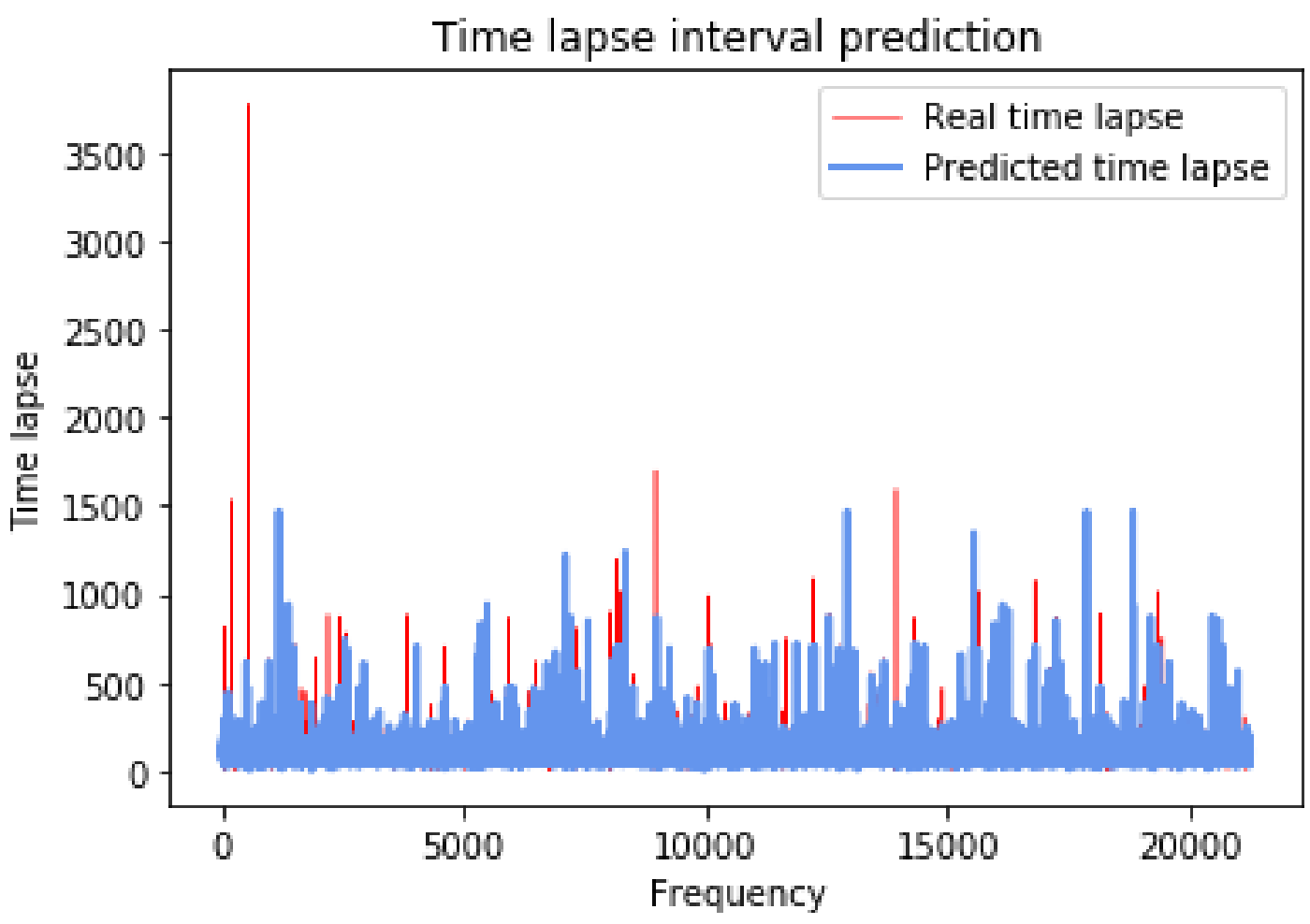

**Fig. 6.** Plot showing the predicted time lapse over the real time lapse for the Decision tree regression model.

The plot in Fig. 6 does not bare evidence of overfitting or underfitting in the model. The visual representation of the model does not show a general output or fit across an input range or a prediction of outlying data-points. Instead, it shows an accurate reading of new inputs spread evenly across the test set. The model performs very well compared to SVM's and k-NN's and this figure reinforces that.

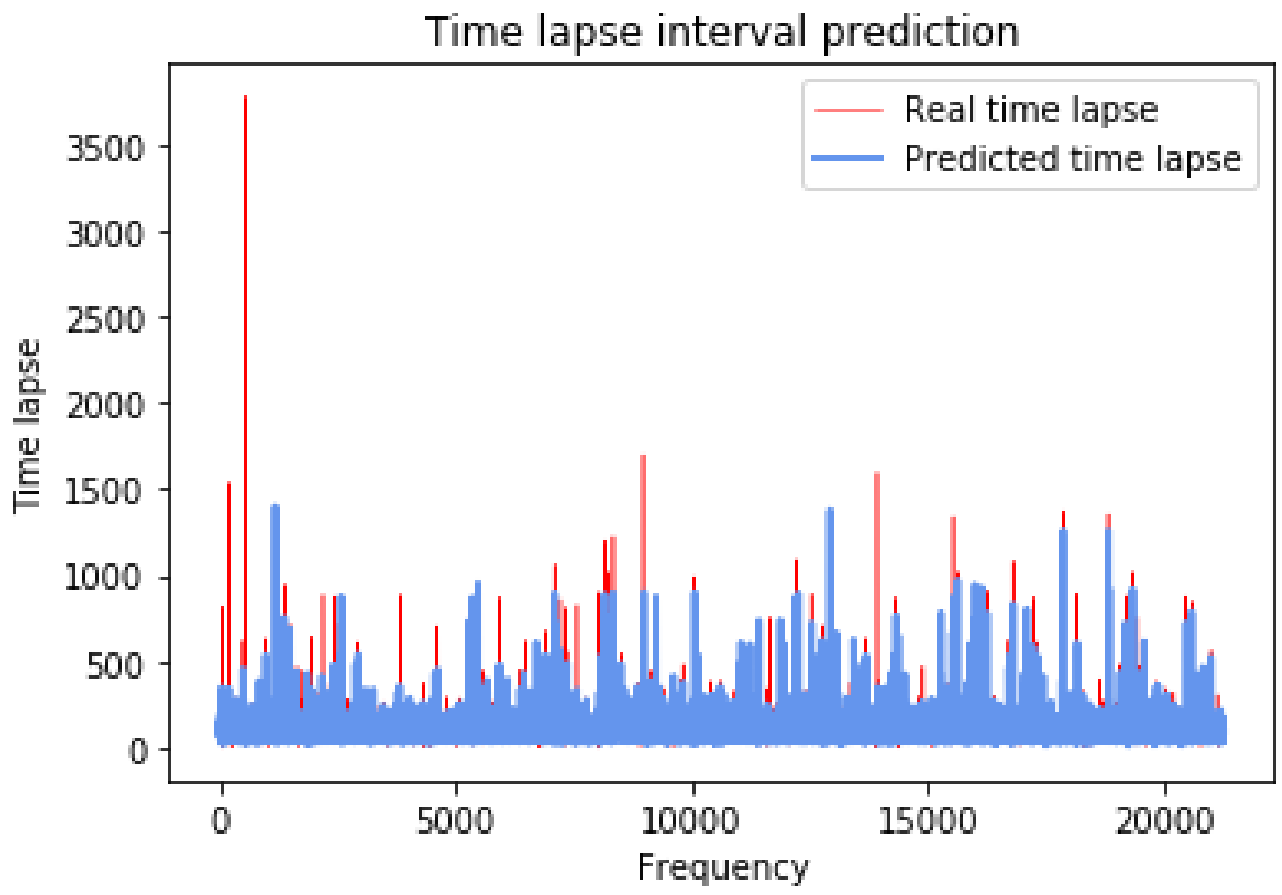

**Fig. 7.** Plot showing the predicted time lapse over the real time lapse for the Decision tree regression model.

The Random Forest plot in Fig. 7 also shows no signs of overfitting or underfitting, on closer inspection, it tends to show a model that does not fit too closely to the data. This calls to mind that the best performing model may not closely fit every data-input, but instead must have a less descriptive analysis when fitting inputs. This is evident with the non-linearity of the data, a less rigorous check on an input from this dataset favors a less descriptive analysis when predicting. Another scenario is random forest being the average of decision trees trains on a range of scores from many decision trees. This normalization on its descriptive analysis may be the reason for this permissible model.

## 5 Discussion

The decision tree cross validation score of 0.66 and accuracy of 67% prediction on the test set shown in Table 5 and Table 6, is a fairly good results but they stand out when compared to the performance of the two other dissimilar models (SVMs and k-NNs). The data used in this experiment is highly non-linear, this means that a unique instance of a feature (i.e, Drug) could map to more than one unique instances of other features. This similarity in the dataset depresses the search space of predictions, models would then act with very minimal discriminations on new inputs. Although the 'Charges sent' feature may have a unique instance across all samples, it still doesn't produce a big enough discrepancy for classification. Therefore, models like SVMs and k-NNs suffer greatly for this, because they assess new inputs on how close they are to a cluster of points(SVMs) or they make an intuitive prediction based on the likeness to other data-points (k-NNs), most of the predictions would then be fairly wrong.
Decision Trees and Random Forests on the other hand define the attention that should be placed on features in new inputs. A new input to the model traverses a tree that classifies based on information and not on distance or likelihood to other data-points. This behavior does not suffer greatly from overly similar or intertwined data instances but bears the risk of overfitting. Thus, a search method was employed to monitor this unwanted attribute in our model. Grid Search, grid search was able to find an optimal depth for the decision tree and thereby improving the performance. The only model that performed better than decision trees is Random Forest, which intuitively makes sense and is expected. Random Forest is an average of a number of these well performing decision trees on predictions and so should have better instances to make a decision from.

## 6 Conclusion

The comparison between these algorithms underscores that a connectivity between unique data-points becomes a major factor when finding a conclusion on suitable machine learning models to employ. In our case, although the goodness of fit scores selected the Random forest and the Decision tree algorithms as optimal, however, the data took a form that was not high in non-linearity and so fell prey to poor scores from algorithms like SVMs and k-NN's. Yet according to the 'no free lunch theorem', this does not deem the Decision Tree and Random Forest algorithm very suitable for analyzing non-linear data; but instead strongly suggests that using k-NN's and SVM's to predict classes betters an application for predictions on well-defined output labels (i.e, Time lapse in days). The decision tree model performs better because it makes a better prediction on new data points than the ballpark estimates made from likelihood models- SVM's and k-NN's. This study also stands as a reference for the importance of grid search when running non-parametric models. Owing to the fact that nonparametric algorithms learn objective functions that are not defined by fixed parameters, grid search presents an appropriate technique that solves for the intricate measure that data needs to be measure with.